\documentclass[10pt,twocolumn,letterpaper]{article}

 \usepackage[pagenumbers]{iccv} 

\definecolor{iccvblue}{rgb}{0.21,0.49,0.74}
\usepackage[pagebackref,breaklinks,colorlinks,allcolors=iccvblue]{hyperref}
\usepackage{array}
\usepackage{multirow}
\usepackage{subcaption}
\usepackage{multirow}
\usepackage{amsmath}
\usepackage{comment}
\usepackage{colortbl}
\definecolor{darkgreen}{rgb}{0.25, 0.5, 0.25}
\usepackage{stfloats}
\usepackage[accsupp]{axessibility}
\def\paperID{10} 
\def\confName{ICCV}
\def\confYear{2025}

\title{MedShift: Implicit Conditional Transport for X-Ray Domain Adaptation}

\author{
Francisco Caetano$^{1,*}$, 
Christiaan Viviers$^{1,*}$, 
Peter H.N. De~With$^{1}$, 
Fons van der Sommen$^{1}$ \\
$^{1}$Eindhoven University of Technology, The Netherlands \\
$^{*}$Equal contribution
}
\begin{document}
\maketitle
\begin{abstract}
Synthetic medical data offers a scalable solution for training robust models, but significant domain gaps limit its generalizability to real-world clinical settings. This paper addresses the challenge of cross-domain translation between synthetic and real X-ray images of the head, focusing on bridging discrepancies in attenuation behavior, noise characteristics, and soft tissue representation. We propose MedShift, a unified class-conditional generative model based on Flow Matching and Schrödinger Bridges, which enables high-fidelity, unpaired image translation across multiple domains. Unlike prior approaches that require domain-specific training or rely on paired data, MedShift learns a shared domain-agnostic latent space and supports seamless translation between any pair of domains seen during training. We introduce X-DigiSkull, a new dataset comprising aligned synthetic and real skull X-rays under varying radiation doses, to benchmark domain translation models. Experimental results demonstrate that, despite its smaller model size compared to diffusion-based approaches, MedShift offers strong performance and remains flexible at inference time, as it can be tuned to prioritize either perceptual fidelity or structural consistency, making it a scalable and generalizable solution for domain adaptation in medical imaging. The code and dataset are available at \href{https://caetas.github.io/medshift.html}{caetas.github.io/medshift.html}.

\end{abstract}

\section{Introduction}

Translating medical images across domains, such as synthetic to real scans, is a key enabler for applications such as clinician training, cross-center harmonization, and domain-robust model development. Differences in imaging protocols, hardware, and data distributions, as well as the often simplified or constrained nature of simulation environments, can lead to substantial domain gaps, reducing the effectiveness of models trained on one dataset when deployed in another. Image-to-image translation offers a solution by mapping images across domains while preserving vital anatomical structural content.

Although paired datasets are rarely available in medical imaging, unpaired translation has been explored primarily through two paradigms: neural style transfer and generative modeling. Neural style-transfer methods rely on matching feature statistics between source and target images, often using perceptual losses and pre-trained networks, but cannot typically model complex structural changes~\cite{jing2019neural}. In contrast, generative approaches, including generative adversarial networks (GANs)\cite{zhu2017unpaired,choi2018stargan}, normalizing flows~(NFs)\cite{grover2020alignflow}, and more recently, diffusion models~(DDPMs)~\cite{benigmim2023one,ozbey2023unsupervised}, learn to transform distributions between domains with greater flexibility and fidelity. However, these models often require training a separate instance for each pair of domains and cannot generalize across multiple domains within a single unified framework.

The goal of this paper is to enable cross-domain generalization between synthetic and real X-ray images of the head, with a specific focus on adapting the appearance and attenuation characteristics of simulated X-ray images to match those observed in real clinical imaging. In practice, this involves learning a domain-transfer model, for example, through style transfer, adversarial training, or feature alignment, that can map or translate synthetic images into the domain of real X-ray images. The key challenge lies in bridging the domain gap, which is primarily caused by differences in: a)~X-ray attenuation profiles, since simulated images may not fully replicate the complex attenuation behavior of X-rays through heterogeneous anatomical structures such as bone, air cavities, and soft tissue; b)~noise characteristics, as real X-ray systems introduce structured noise, scatter, and compression artifacts that are not present in simulators; and c)~soft-tissue representation and contrast dynamics, particularly at the boundaries of bone or within overlapping anatomical features.

\begin{figure*}[ht]
    \centering
    \includegraphics[width=\linewidth]{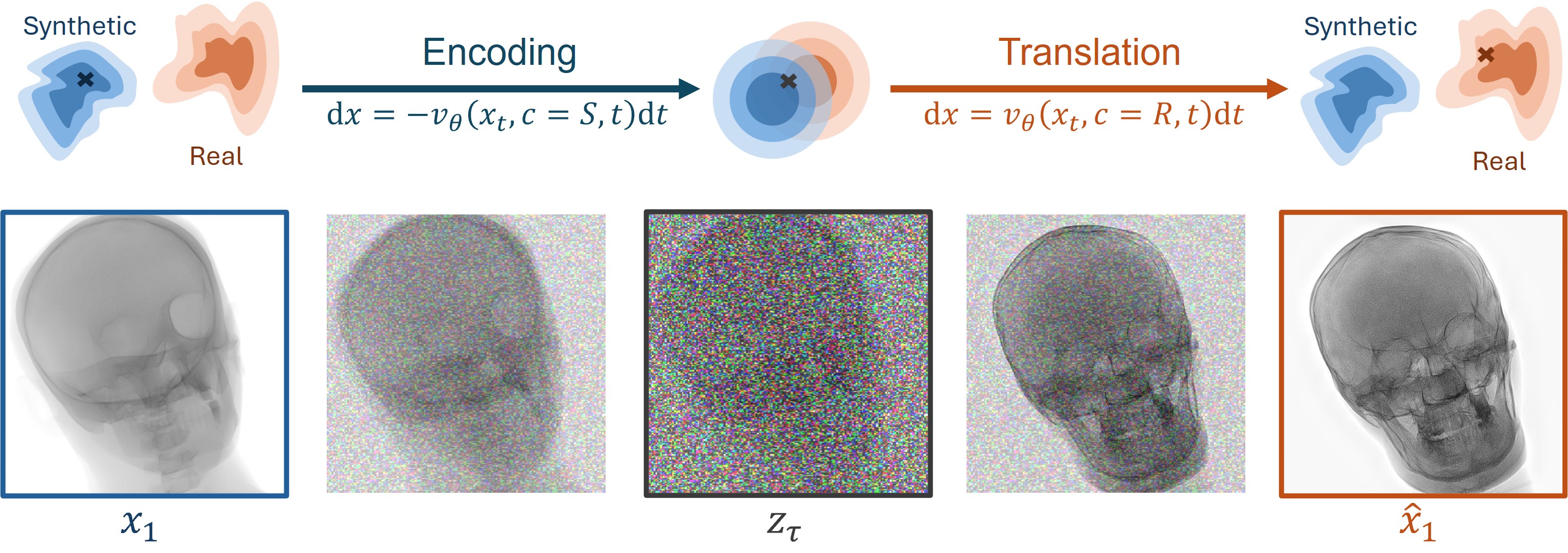}
    \caption{Overview of MedShift inference. A source image $x_1$ is first \textcolor{blue}{\textbf{encoded}} into a domain-agnostic latent representation $z_{\tau}$. This latent lies near a shared manifold across all domains. Then, \textcolor{orange}{\textbf{translation}} is performed by forward-time sampling conditioned on the target domain label to obtain the translated image $\hat{x}_1$.}
    \label{fig:medshift}
\end{figure*}

By learning this adaptation, the system aims to capture how real bone attenuates X-ray signals, including nuances such as cortical thickness, trabecular density, and beam hardening effects, which are often oversimplified in simulators. This enables models trained or tested on synthetic data to become more predictive, reliable, and generalizable in real-world applications, such as in surgical navigation, training, or image-guided interventions.

In this work, we propose MedShift, a unified conditional generative model to translate medical images across multiple domains. MedShift learns a single implicit transport map conditioned on domain labels, leveraging flow matching and optimal transport to align source and target distributions. Crucially, the model supports standard sampling and enables translation between any pair of domains seen during training. Our contributions are threefold: 1)~we introduce a novel model based on Flow Matching and Schrödinger Bridges for unpaired image translation at high resolution; 2)~we release X-DigiSkull, a new dataset of synthetic and real X-ray scans, including multi-angle acquisitions and varying radiation dosages; and 3)~we benchmark several state-of-the-art generative models on this dataset, establishing a reference for future work.
\section{Related Work}

\subsection{Neural Style Transfer}

Neural style transfer has undergone rapid advancement, beginning with the seminal work of Gatys~\etal~\cite{gatys2016image}, who introduced an optimization-based approach leveraging pretrained convolutional neural networks to disentangle and recombine content and style. Despite producing high-quality results, such methods are computationally intensive and constrained in flexibility. To address these limitations, later work explored feedforward architectures~\cite{ulyanov2016texture} and GAN-based models~\cite{chen2018cartoongan}, allowing real-time stylization and improved scalability. More recently, attention mechanisms and Transformer-based architectures~\cite{yao2019attention} have been used to enhance spatial correspondence and semantic alignment, pushing the boundaries of visual fidelity and control, albeit with increased model complexity and computational cost.

\subsection{Image-to-Image Translation}

Recent advances in generative models have driven progress in image-to-image translation. Paired methods~\cite{zhao2021large,zhu2020sean} rely on supervised training with reconstruction and adversarial losses, but require aligned datasets. Conditional diffusion models have extended this space, incorporating text or spatial conditioning and building on large pretrained models such as GLIGEN~\cite{li2023gligen}, T2I-Adapter~\cite{mou2024t2i}, and ControlNet~\cite{zhang2023adding}. However, they still depend on paired data. In unpaired settings, common strategies include cycle consistency~\cite{zhu2017unpaired}, shared latent spaces~\cite{lee2018diverse}, content preservation~\cite{shrivastava2017learning}, and contrastive learning~\cite{han2021dual}. Recent work has explored unpaired diffusion models~\cite{sasaki2021unit}, but this approach typically requires domain-specific training from scratch. In contrast, CycleGAN-Turbo~\cite{parmar2024one} leverages a pretrained diffusion model in a CycleGAN~\cite{zhu2020unpairedimagetoimagetranslationusing} framework for unpaired translation, eliminating the need for paired data while allowing faster and more scalable inference.

\begin{figure*}[!t]
    \centering
    \includegraphics[width=\linewidth]{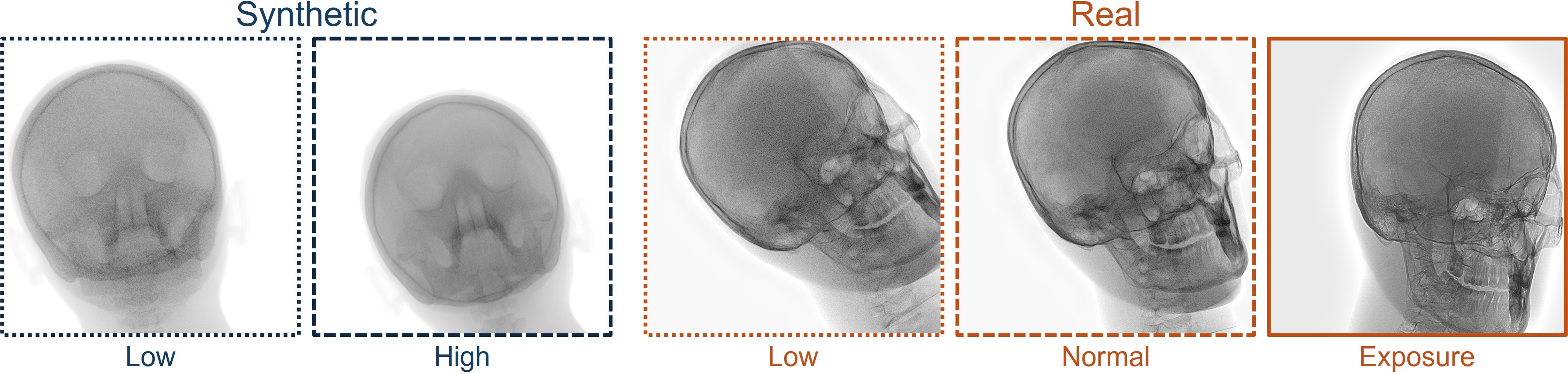}
    \caption{Dataset overview. The \textcolor{blue}{synthetic} domain contains \textit{Low} and \textit{High} dosage samples generated using the Mentice VIST$\textsuperscript{\textregistered}$ simulator; the \textcolor{orange}{real} domain includes \textit{Low}, \textit{Normal}, and \textit{Exposure} dosage categories acquired from a skull phantom using the Philips Azurion IGT system.}
    \label{fig:medshift_data}
\end{figure*}

\subsection{Score-based Models \& Flow Matching}

Score-based generative models~\cite{song2020score,song2020improved} learn the gradient of the data distribution, known as the score function, enabling sample generation through stochastic processes such as Langevin dynamics or Stochastic Differential Equations. Although effective, these models are often computationally expensive due to the need for iterative sampling over many steps. Flow Matching~(FM)~\cite{lipman2022flow,lipman2024flow} addresses this inefficiency by directly parameterizing a time-dependent vector field that deterministically transports samples from a known base distribution to the target data distribution via an ordinary differential equation~(ODE). By supervising intermediate steps using closed-form conditional distributions, FM offers a scalable and efficient alternative to traditional diffusion-based methods, without sacrificing generative quality.

\subsection{Schrödinger Bridges}

Score-based generative models have been shown to approximate Schrödinger bridges, which model the most likely stochastic path that connects two marginal distributions~\cite{chen2021likelihood}. This connection enables image translation as a form of marginal-matching interpolation. Dual Diffusion Implicit Bridges~\cite{su2022dual} leverage this by using two deterministic ODEs, parameterized by separate diffusion models trained in each domain, to map a source image to a latent code and decode it into the target domain. Cycle Diffusion~\cite{wu2023latent} extends this idea further, demonstrating that such mappings can be performed within a single model in latent space.
\section{MedShift}
\label{sec:medshift}

MedShift is a class-conditional Flow Matching model for high-resolution image translation across X-Ray domains. Each class corresponds to a unique domain, such as simulated vs. real X-rays, which will be referred to as class $S$ and class $R$, respectively. The model is trained using classifier-free guidance~(CFG), allowing it to learn conditional score estimates for each domain without relying on paired data. MedShift operates in a latent space learned through a pretrained VAE, significantly reducing computational cost while preserving semantic detail and spatial resolution.

During inference, a source image, such as a simulated X-ray (class S), is first encoded into a domain-agnostic latent representation via time integration. Starting from the observed image $x_1$, we integrate backward from $t = 1$ to an intermediate time $\tau \in (0, 1)$ under the source domain condition $c = S$, yielding the latent representation $z_\tau$:
\begin{equation}
    z_\tau = x_1 - \int_\tau^1 v_\theta(x_t, c = S, t)\, dt.
\end{equation}
This intermediate state lies in a shared latent manifold that is approximately aligned across all domains, as shown in Section~\ref{sec:latent}. To generate the translated image in a target domain, e.g., a real X-ray at high dose, we then integrate forward from $\tau$ to $1$, this time conditioning the target domain $c = R$:
\begin{equation}
    \hat{x}_1 = z_\tau + \int_\tau^1 v_\theta(x_t, c = R, t)\, dt.
\end{equation}
This two-stage process, consisting of \textcolor{blue}{\textbf{encoding}} and \textcolor{orange}{\textbf{translation}}, enables faithful domain transfer while preserving essential anatomical content. Figure~\ref{fig:medshift} illustrates this conditional transport mechanism between domains.
\section{Methodology}

\begin{table}[!b]
    \centering
    \caption{Dataset statistics showing the number of training and test images across domains and dosage levels.}
    \label{tab:dataset}
    \resizebox{\linewidth}{!}{%
    \begin{tabular}{ll|c|c|c}
        \toprule
        \textbf{Domain} & \textbf{Dosage} & \textbf{Training Images} & \textbf{Test Images} & \textbf{Total Images} \\
        \midrule
        \multirow{2}{*}{Synthetic}
        & Low & 4,979 & 853 & 5,832 \\
        & High & 4,979 & 853 & 5,832 \\
        \midrule
        \multirow{3}{*}{Real}
        & Low & 1,857 & 330 & 2,187 \\
        & Normal & 1,857 & 330 & 2,187 \\
        & Exposure & 1,853 & 329 & 2,182 \\
        \bottomrule
    \end{tabular}}
    \end{table}

\begin{table*}[!t]
    \centering
    \caption{Comparison of style transfer outputs across the five methods. All the outputs are derived from the synthetic inputs.}
    \label{tab:qualitative}
    \resizebox{\textwidth}{!}{%
    \begin{tabular}{>{\centering\arraybackslash}m{0.142\textwidth} >{\centering\arraybackslash}m{0.142\textwidth} >{\centering\arraybackslash}m{0.142\textwidth} >{\centering\arraybackslash}m{0.142\textwidth} >{\centering\arraybackslash}m{0.142\textwidth} >{\centering\arraybackslash}m{0.142\textwidth} >{\centering\arraybackslash}m{0.142\textwidth}}
        \toprule
        \textbf{Synthetic} & \textbf{Closest Real} & \textbf{Hierarchy Flow (st=0.25)} & \textbf{CycleGAN (ss=0.0)} & \textbf{Z-STAR} & \textbf{SDEdit (st=0.2)} & \textbf{MedShift ($\boldsymbol{\tau}$=0.45)} \\
        \midrule
        \includegraphics[width=\linewidth]{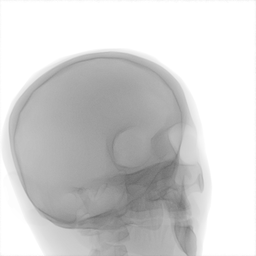} &
        \includegraphics[width=\linewidth]{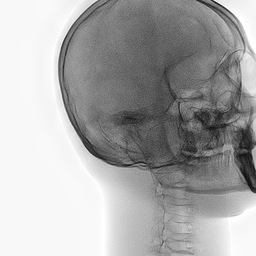} &
        \includegraphics[width=\linewidth]{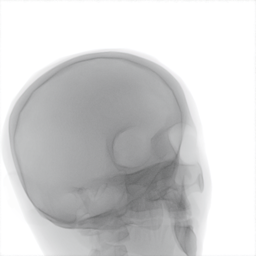} &
        \includegraphics[width=\linewidth]{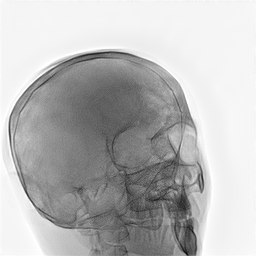} &
        \includegraphics[width=\linewidth]{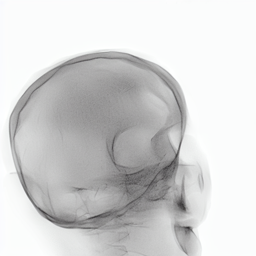} &
        \includegraphics[width=\linewidth]{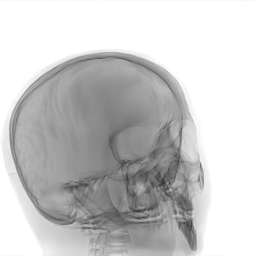} &
        \includegraphics[width=\linewidth]{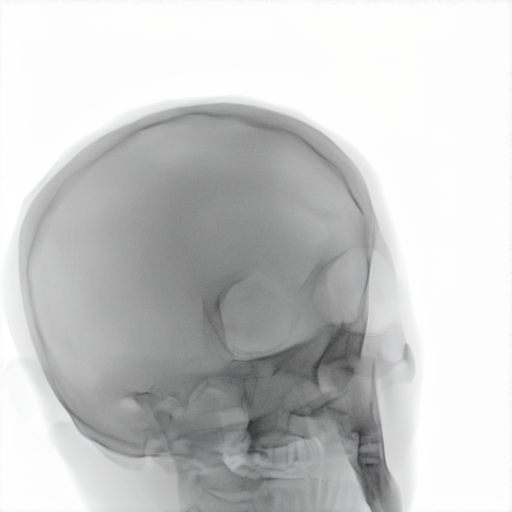} \\
        
        \includegraphics[width=\linewidth]{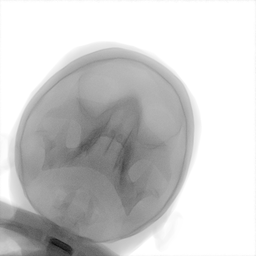} &
        \includegraphics[width=\linewidth]{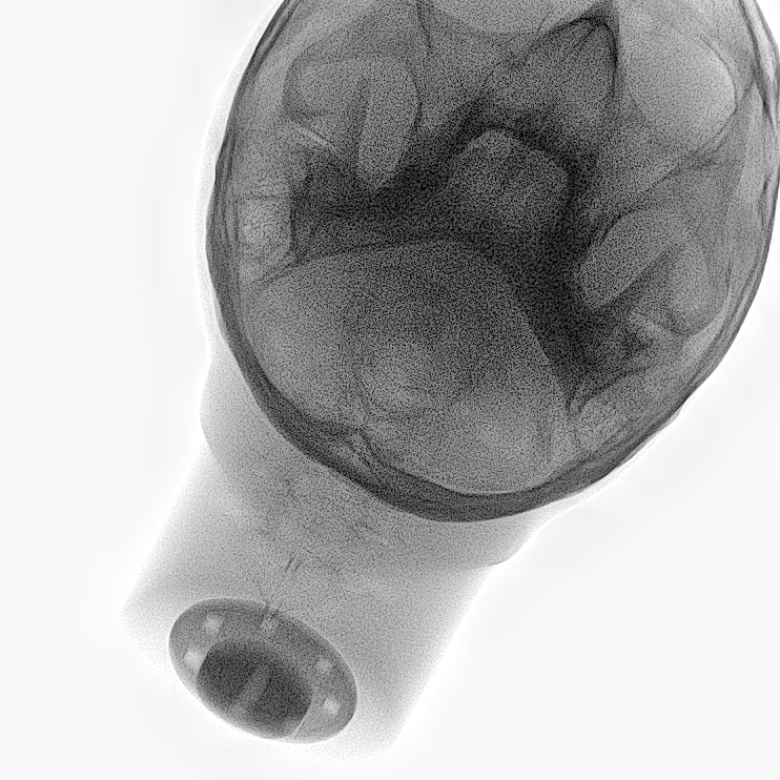} &
        \includegraphics[width=\linewidth]{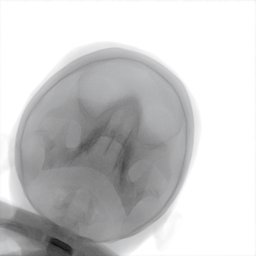} &
        \includegraphics[width=\linewidth]{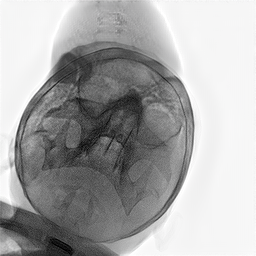} &
        \includegraphics[width=\linewidth]{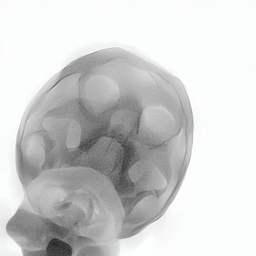} &
        \includegraphics[width=\linewidth]{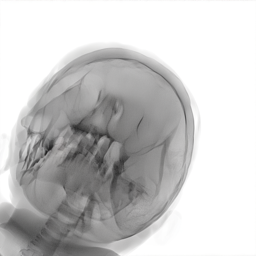} &
        \includegraphics[width=\linewidth]{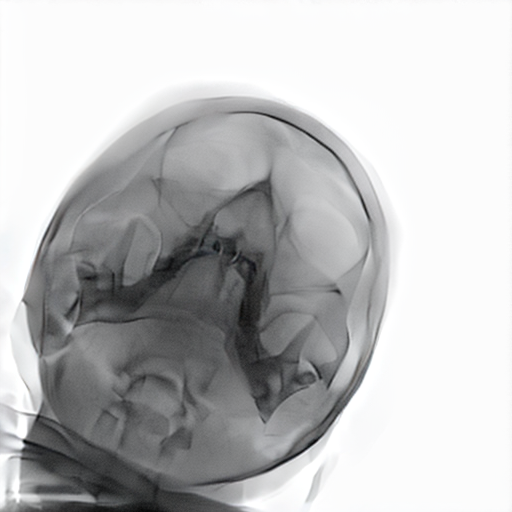} \\

        \includegraphics[width=\linewidth]{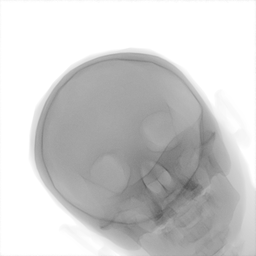} &
        \includegraphics[width=\linewidth]{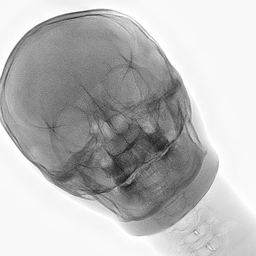} &
        \includegraphics[width=\linewidth]{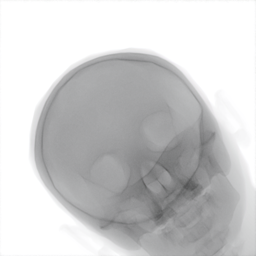} &
        \includegraphics[width=\linewidth]{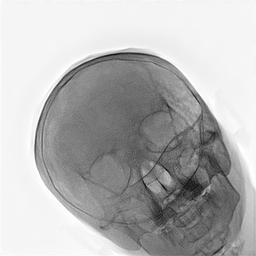} &
        \includegraphics[width=\linewidth]{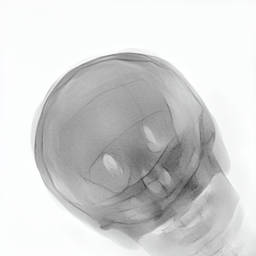} &
        \includegraphics[width=\linewidth]{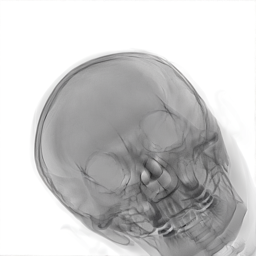} &
        \includegraphics[width=\linewidth]{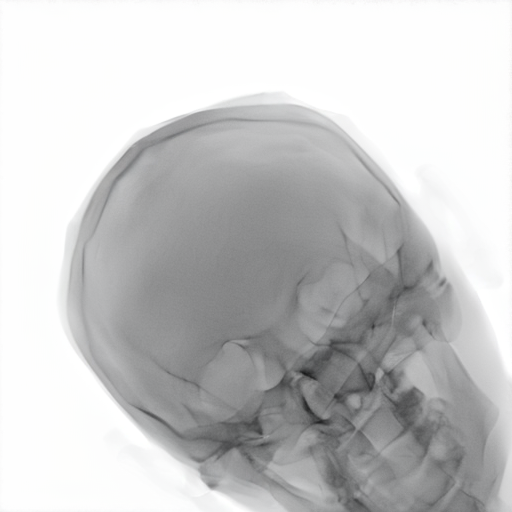} \\
        \bottomrule
    \end{tabular}}
\end{table*}

\subsection{Datasets}


We develop and release a new dataset of real and synthetic head X-ray images, X-DigiSkull, to study the domain adaptation. The dataset consists of synthetic X-ray images of a human skull generated using the Mentice VIST$\textsuperscript{\textregistered}$ simulator~\footnote{\href{https://www.mentice.com/simulator/vist-g7}{https://www.mentice.com/simulator/vist-g7}}. Real‑time X‑rays are generated by holding the 3D voxel “patient” head model with per‑voxel attenuation, casting one ray per detector pixel to form a digitally reconstructed radiograph and then approximating scatter, focal‑spot and detector blur, grid/heel effects and detector response, adding quantum/read noise and final image post-processing as the C‑arm and devices move. Real images are acquired from a clinical-grade physical skull phantom using the Philips Azurion Image Guided Therapy~(IGT) system. Images are captured from common IGT working positions for neuro procedures. The dataset consists of multiple orientations and is available in three different radiation dose settings: \emph{low}, \emph{normal}, and \emph{exposure}~(Philips exclusive), the latter offering enhanced image quality and detail, as shown in Figure~\ref{fig:medshift_data}. This consists of viewing angles $r_{z} \in  [-40^{\circ},+40^{\circ}]$, $r_{y} \in  [-40^{\circ},+40^{\circ}]$, $r_{x} \in [-40^{\circ},+40^{\circ}]$ with respect to the starting position in 10$^\circ$ increments and up to 3 images at each position to capture the noise present. This results in a total of 2,187 real images. The coordinate systems of the real and synthetic environment are aligned and synthetic images are rendered to approximate the same viewpoints as the real phantom images with the patient table starting at a similar position.The head 3D model used in the simulation is meticulously built from a real clinical case. We capture the synthetic images in finer increments of 5$^\circ$ across the three angles, producing the 5,832 ($18^3$) images~\footnote{Data available at \href{https://zenodo.org/records/16535437}{https://zenodo.org/records/16535437}}. The aim of this alignment is not to achieve precise supervised image translation, but rather to establish a consistent reference structure that preserves spatial features. The composition of the dataset is summarized in Table~\ref{tab:dataset}, which contains information on the splits and the number of images available. All images are cropped and resized to 780$\times$780~pixels. The test set is obtained by uniformly sampling 15\% of viewing angles and corresponding images to ensure a representative distribution. For our experiments, we focus on the task of converting synthetic images at \emph{high} dose to real images at \emph{normal} dose.

\subsection{Models}

The proposed method is compared with recent generative models, including GANs, NFs, and DDPMs.

\subsubsection{Hierarchy Flow}
\emph{Hierarchy Flow}~(HF)~\cite{fan2023hierarchy} is trained as an unpaired domain adapter. In each step, a synthetic X-ray serves as the content image, while an approximately aligned X-ray provides style. The synthetic image is encoded by a stack of hierarchical coupling layers, while the real image passes through a style encoder that outputs per-channel means and variances. Adaptive Instance Normalization swaps these statistics into the encoded synthetic features, and the invertible network is run backward to reconstruct a candidate real-looking radiograph. Training minimizes a content loss between the output and the synthetic input and an alignment-style loss that matches only the most semantically relevant feature channels to the real sample. Different strengths of the style loss (st) are experimented with to enforce more target domain style onto the synthetic image.

\subsection{CycleGAN-Turbo}
We adapt \emph{CycleGAN-Turbo}~\cite{parmar2024one}, a one-step variant of Stable Diffusion~Turbo, to unpaired synthetic to real skull X-ray translation. Each training batch pairs a synthetic image and the corresponding prompt ``high-dose \textit{SYNTHETIC} X-ray image of a human skull'' with random real images labeled ``normal-dose \textit{REAL} X-ray image of a human skull.''  The image is fed into the Stable Diffusion pipeline, in which only the first convolution, LoRA adapters, and zero-convolution skip connections are trained. The model is optimized with a CycleGAN objective: a CLIP-guided discriminator enforces adversarial alignment to the target domain; a cycle-consistency loss (LPIPS~+~$L_1$) preserves structure; and an identity loss stabilizes intensity and dose. Along with a large hyperparameter search, we conduct an additional experiment where we investigate adding a structural similarity loss between the input and generated images to attempt to enforce further structural coherence.

\subsubsection{SDEdit}
Following \emph{SDEdit}~\cite{mengsdedit}, we fully fine-tune a pretrained Stable Diffusion 2.1 network in both domains, annotating the source images with the prompt ``high-dose \textit{SYNTHETIC} X-ray image of a human skull'' and the target images with ``normal-dose \textit{REAL} X-ray image of a human skull''. At test time, a synthetic image is diffused to a user-chosen noise level $st\in[0,1]$ and then reconstructed by reverse SDE while conditioned on the real domain prompt. This denoising step injects the appearance statistics of real X-rays yet preserves the anatomical structure of the input, attempting to produce a realistic high-dose skull radiograph without requiring paired supervision.

\subsubsection{Z-STAR}
\emph{Z-STAR}~\cite{deng2023z}, a training-free attention-rearrangement strategy, is also used. A synthetic skull X-ray $I_{\text{syn}}$ provides \emph{content}, whereas an approximately aligned real X-ray $I_{\text{real}}$ provides \emph{style}. Both images are inverted through a DDIM solver, giving latent paths $x^{c}_{0:T}$ and $x^{s}_{0:T}$. During reverse diffusion, we rearrange the U-Net cross-attention: queries are taken from $x^{c}$, while keys/values stack $\{x^{c},x^{s}\}$ and are jointly normalized, suppressing style tokens that poorly match the content. The resulting attention guides denoising to inject real-domain dose and texture while preserving skull structure, producing realistic high-dose radiographs with no additional training or supervision.

\subsubsection{MedShift}
\emph{MedShift} is evaluated using an Euler ODE solver taking 50 integration steps; the CFG scale is set to 8.5. The hyperparameters and training setup are listed in Appendix~\ref{sec:implementation_sup}.

\subsection{Metrics}

We evaluate the quality of domain-transferred images using a combination of realism and structure preservation metrics. CFID~(Conditional Fréchet Inception Distance), Coverage, and CMMD~(Conditional Maximum Mean Discrepancy) evaluate how well the generated samples match the distribution of the target domain. CFID captures global alignment in feature space, Coverage quantifies the proportion of the target distribution covered by generated samples, and CMMD measures discrepancies between conditional distributions. To assess whether the anatomical structure is preserved, we report on LPIPS~(Learning Perceptual Image Patch Similarity) and SSIM~(Structural Similarity Index Measure). LPIPS compares deep feature similarity between source and generated images, while SSIM evaluates structural similarity. For SDEdit and MedShift, the results are reported as mean and a confidence interval of two standard deviations measured over three independent runs and checkpoints, respectively. Optimal performance is not achieved by strictly maximizing similarity to the target domain or by rigidly preserving the source structure. Instead, it requires a balanced trade-off between both objectives. To capture this, we rank models separately on realism and structure preservation and report the average rank as a measure of overall performance.
\section{Results \& Discussion}

\subsection{Benchmark}
\begin{figure}[b]
    \centering
    \includegraphics[width=\linewidth]{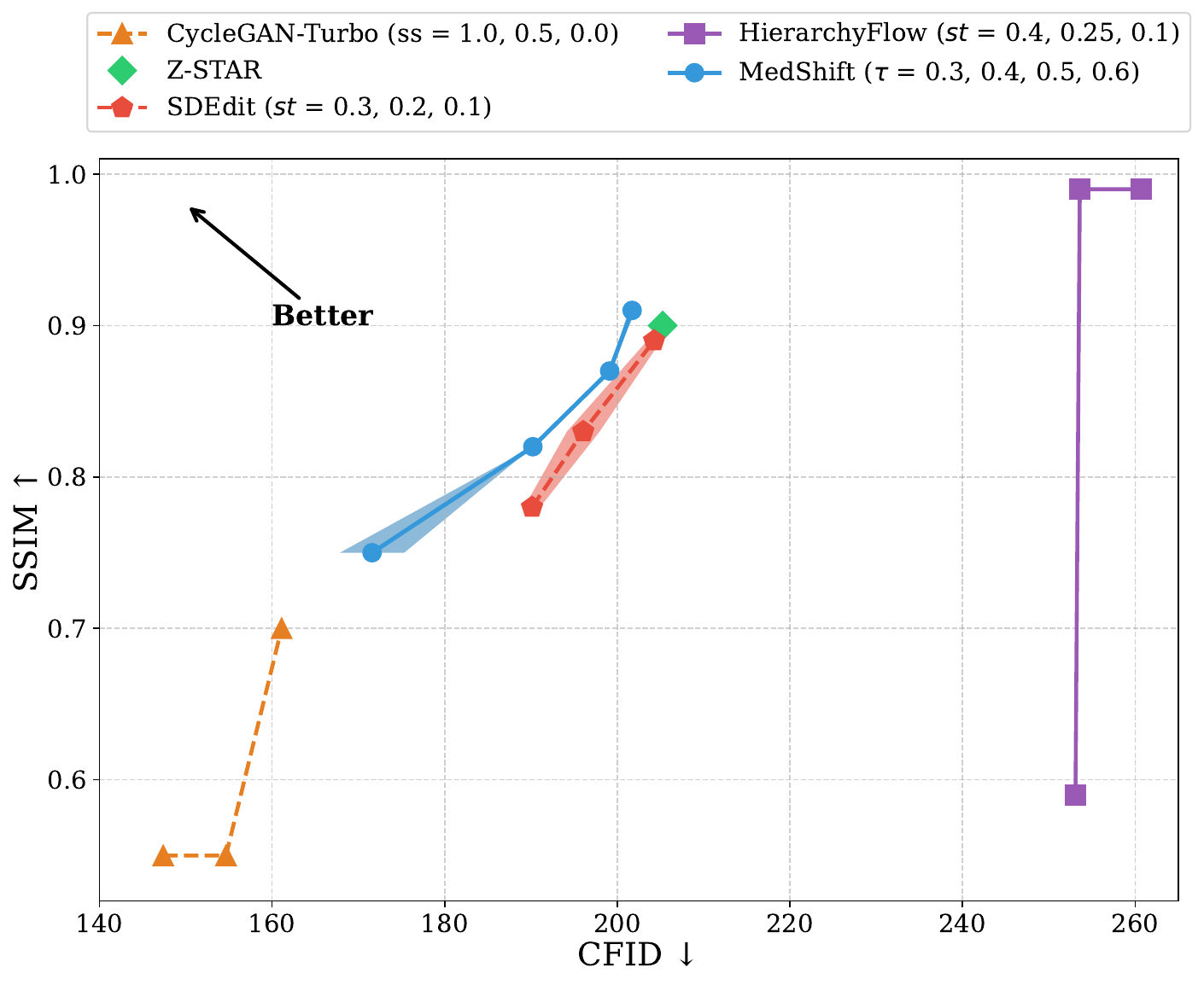}
    \caption{Trade-off between structural fidelity~(SSIM) and realism~(CFID) for the evaluated models.}
    \label{fig:tradeoff}
\end{figure}
\begin{table*}[t]
    \centering
    \caption{Comparison of image translation performance of the methods. Global performance is evaluated through the \textcolor{darkgreen}{average rank}.}
    \label{tab:results}
    \resizebox{\textwidth}{!}{%
    \begin{tabular}{ll|c|c|c|c|c|c|c|c}
        \toprule
         \multirow{2}{*}{\textbf{Type}} & \multirow{2}{*}{\textbf{Method}} & \multicolumn{4}{c|}{\textbf{Realism}} & \multicolumn{3}{c|}{\textbf{Structure}} & \textbf{Average} \\
         & & \textbf{CFID}($\downarrow$) & \textbf{Cov.}($\uparrow$) & \textbf{CMMD}($\downarrow$) & \textbf{Rank}($\downarrow$) & \textbf{LPIPS}($\downarrow$) & \textbf{SSIM}($\uparrow$) & \textbf{Rank}($\downarrow$) & \textbf{Rank}($\downarrow$) \\
         \midrule
         None & Synthetic Images & 262.56 & 0.48 & 10.46 & -- & 0.00 & 1.00 & -- & --\\
         \midrule
          \multirow{3}{*}{NF} & Hierarchy Flow~\cite{fan2023hierarchy}~(st$=$0.1) & 260.75 & 0.47 & 10.62 & 13 & 0.01 & 0.99 & \textbf{1} & \cellcolor{gray!10!green!10}7\\
          & Hierarchy Flow~\cite{fan2023hierarchy}~(st$=$0.25) & 253.59 & 0.50 & 10.64 & 12 & 0.01 & 0.99 & \textbf{1} & \cellcolor{gray!10!green!10}6.5\\
          & Hierarchy Flow~\cite{fan2023hierarchy}~(st$=$0.4) & 253.09 & 0.55 & 12.02 & 11 & 0.40 & 0.58 & \textbf{11} & \cellcolor{gray!10!green!10}11\\
         \midrule
          \multirow{3}{*}{GAN} & CycleGAN-Turbo~\cite{parmar2024one}~(ss=0.0) & 161.11 & 0.85 & 5.68 & \textbf{3} & 0.47 & 0.70 & 10 & \cellcolor{gray!10!green!10}6.5\\
          & CycleGAN-Turbo~\cite{parmar2024one}~(ss=0.5) & 154.66 & 0.81 & 1.89 & \textbf{2} & 0.52 & 0.55 & 13 & \cellcolor{gray!10!green!10}7.5\\
          & CycleGAN-Turbo~\cite{parmar2024one}~(ss=1.0) & 147.39 & 0.86 & 2.51 & \textbf{1} & 0.51 & 0.56 & 12 & \cellcolor{gray!10!green!10}6.5\\
          \midrule
          \multirow{4}{*}{DDPM}& Z-STAR~\cite{deng2023z} & 205.26 & 0.60 & 9.41 & 10 & 0.13 & 0.90 & 4 & \cellcolor{gray!10!green!10}7 \\
          & SDEdit~\cite{mengsdedit}~(st$=$0.1) & 204.27$\pm$0.80 &  0.65$\pm$0.13 & 7.18$\pm$0.1 & 9 & 0.12$\pm$0.00 & 0.89$\pm$0.00 & 5 & \cellcolor{gray!10!green!10}7 \\
          & SDEdit~\cite{mengsdedit}~(st$=$0.2) & 196.06$\pm$1.89 &  0.70$\pm$0.04 & 7.47$\pm$0.05 & 7 & 0.17$\pm$0.00 & 0.83$\pm$0.00 & 7 & \cellcolor{gray!10!green!10}7 \\
          & SDEdit~\cite{mengsdedit}~(st$=$0.3) & 190.12$\pm$0.92 &  0.71$\pm$0.04 & 7.89$\pm$0.09 & 5 & 0.21$\pm$0.00 & 0.78$\pm$0.00 & 8 & \cellcolor{gray!10!green!10}6.5\\
          \midrule
          \multirow{3}{*}{FM} & MedShift~(Prop., $\tau$$=$0.6) & 201.72$\pm$0.41 &  0.65$\pm$0.07  & 8.10$\pm$0.14  & 8  & 0.09$\pm$0.00  & 0.91$\pm$0.00  & 3 & \cellcolor{gray!10!green!10}\textbf{5.5} \\
          & MedShift~(Prop., $\tau$$=$0.45) & 195.17$\pm$0.02  & 0.72$\pm$0.01  & 8.17$\pm$0.01  & 6 & 0.14$\pm$0.00  & 0.85$\pm$0.00  & 6 & \cellcolor{gray!10!green!10}6\\
          & MedShift~(Prop., $\tau$$=$0.3) & 171.59$\pm$3.72   & 0.71$\pm$0.01  & 8.14$\pm$0.26  & 4  & 0.24$\pm$0.00  & 0.75$\pm$0.00  & 9 & \cellcolor{gray!10!green!10}6.5\\
        \bottomrule
    \end{tabular}}
\end{table*}
As shown in Table~\ref{tab:results}, CycleGAN-Turbo achieves the strongest performance on distributional metrics such as CFID and density, but this comes at the cost of anatomical fidelity. The implementation of the structural similarity loss did not yield the expected outcomes. The qualitative results in Table~\ref{tab:qualitative} reveal that it introduces spurious features, especially in the second row, undermining structural correctness. In contrast, Hierarchy Flow preserves anatomical details exceptionally well by applying minimal transformation, resulting in outputs that closely resemble the input and thus offer limited to no domain adaptation. As noted in Figure~\ref{fig:tradeoff} and Table~\ref{tab:results}, the model completely breaks down at high style strengths values without improvement in FID.

MedShift is evaluated in three $\tau$ settings to explore the trade-off between structural preservation and generative realism. At $\tau=0.6$, the model is second only to HierarchyFlow in maintaining structure, while significantly outperforming it in CFID and coverage. The low-fidelity setting ($\tau=0.3$) reaches CFID values comparable to CycleGAN but with far fewer anatomical distortions. The intermediate configuration ($\tau=0.45$) provides a good trade-off, as seen in Table~\ref{tab:qualitative}. Z-STAR maintains structure but fails to transfer style effectively to the lower jaw, while SDEdit captures pixel intensities well but introduces artifacts into the cranial region, particularly in the second example. As shown in Figure~\ref{fig:tradeoff}, MedShift achieves a more favorable balance between structural fidelity and image realism across all $\tau$ settings, outperforming the other models in this trade-off space. This reinforces the findings based on average ranking, confirming MedShift as the most well-rounded strategy.

An important architectural distinction of MedShift is its memory efficiency. Unlike the diffusion-based models, which incorporate a Stable Diffusion U-Net, MedShift leverages a smaller custom U-Net, resulting in significantly enhanced efficiency and reduced training times.

\subsection{Ablation Study}\label{ablation}

We perform an ablation study to assess the impact of the denoising parameter $\tau$ and the CFG scale on image quality (Table~\ref{tab:ablation_tau_cfg}). The results show clear trends: Higher $\tau$ values enhance structural fidelity, while lower values increase stylistic transformation. In contrast, increasing CFG improves alignment with the target domain, but introduces greater deviation from the source anatomy; reducing CFG maintains structural features at the cost of style transfer. In particular, both parameters are inference-time controls, requiring no retraining, and thus offer flexibility to tailor output to specific clinical or application needs. 

Table~\ref{tab:additional_results} shows more qualitative results that highlight how the denoising parameter $\tau$ affects the style transfer process. When $\tau$ is low, the model strongly pushes the images toward the target domain style. In some cases, especially at $\tau=0.3$, this leads to hallucinated structures that do not exist in the original image, similar to what we see with CycleGAN-Turbo. Although these outputs may appear visually plausible in the target domain, they no longer preserve the anatomy of the source, which is crucial in medical imaging.

As $\tau$ increases, the outputs stay closer to the original structure. At midrange values like $\tau=0.5$, we get a good balance: the contrast in darker areas improves, the target style is seen, and the key structures are still intact. But if we go too far towards low $\tau$, some artifacts start to appear, like distorted soft tissue or over-sharpened edges.

\begin{table*}[tbp]
    \centering
    \caption{Comparison of domain transfer outputs across four different values of $\tau$, with CFG=8.5.}
    \resizebox{\textwidth}{!}{%
    \begin{tabular}{c|cccc}
        \toprule
        \textbf{Model} & \textbf{Sample 1} & \textbf{Sample 2} & \textbf{Sample 3} & \textbf{Sample 4}  \\
        \midrule
        \raisebox{0.9cm}{Synthetic} &
        \includegraphics[width=0.18\textwidth]{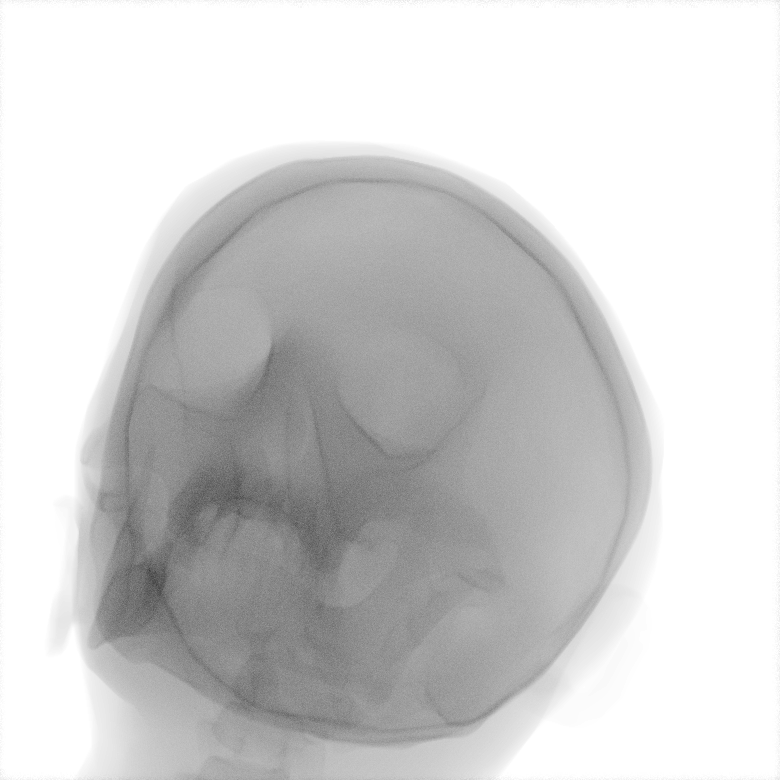} &
        \includegraphics[width=0.18\textwidth]{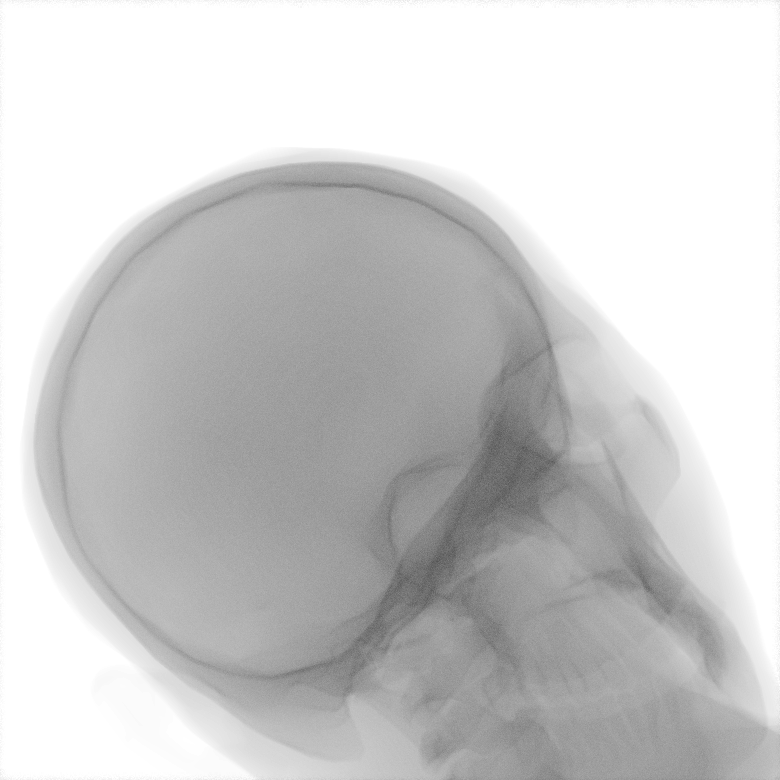} &
        \includegraphics[width=0.18\textwidth]{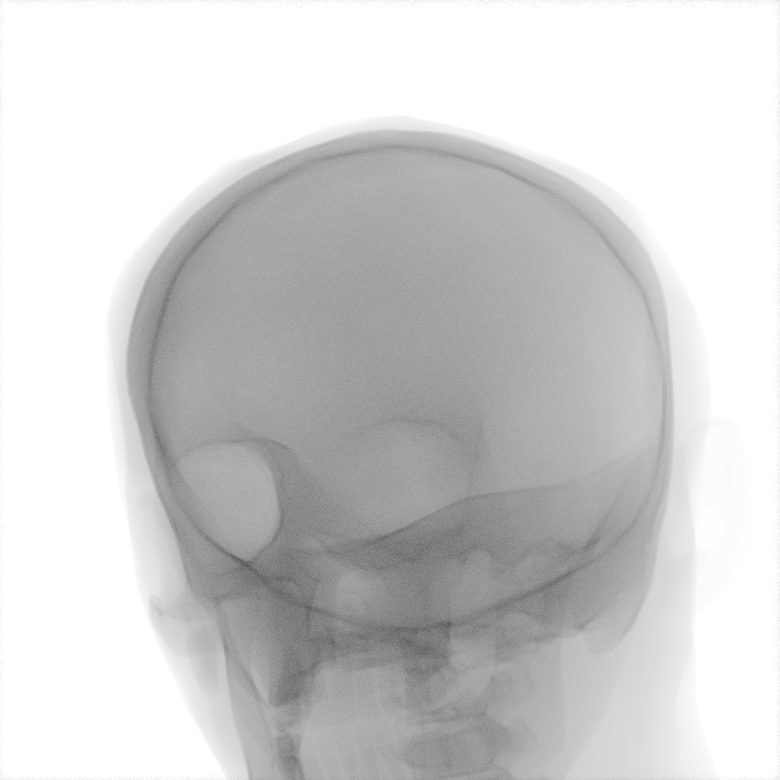} &
        \includegraphics[width=0.18\textwidth]{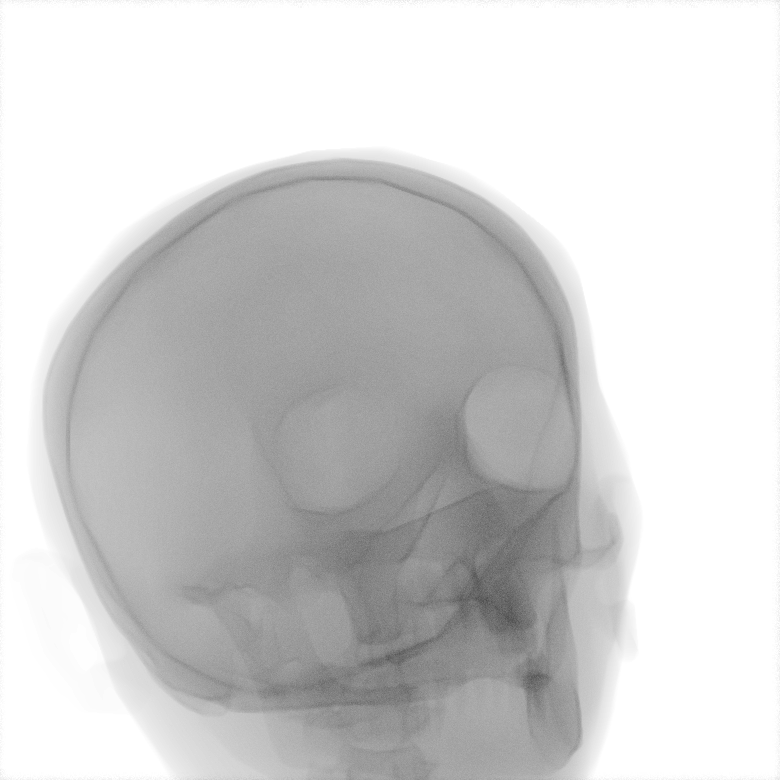} \\
        
        \raisebox{0.9cm}{Closest Real} &
        \includegraphics[width=0.18\textwidth]{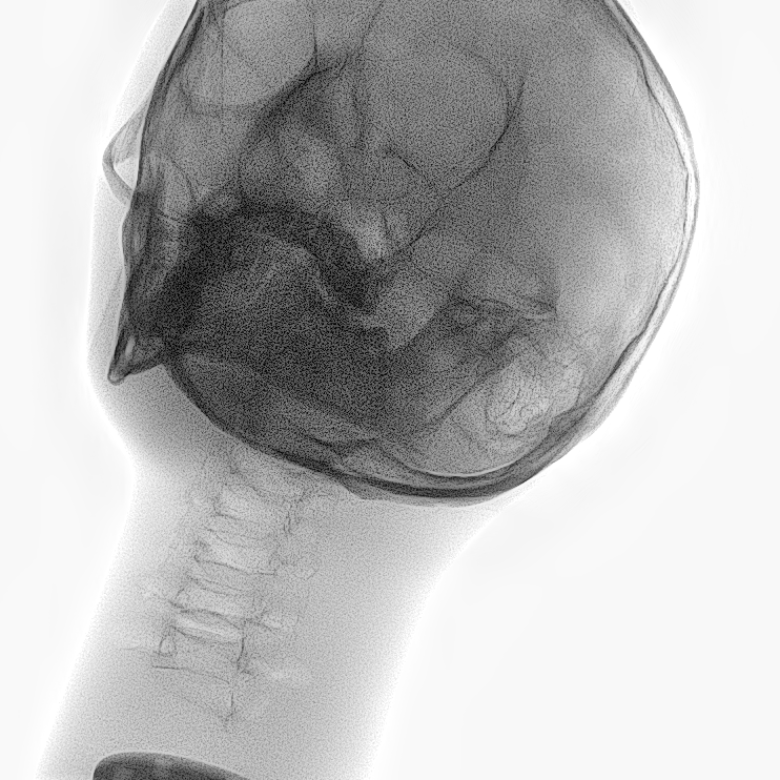} &
        \includegraphics[width=0.18\textwidth]{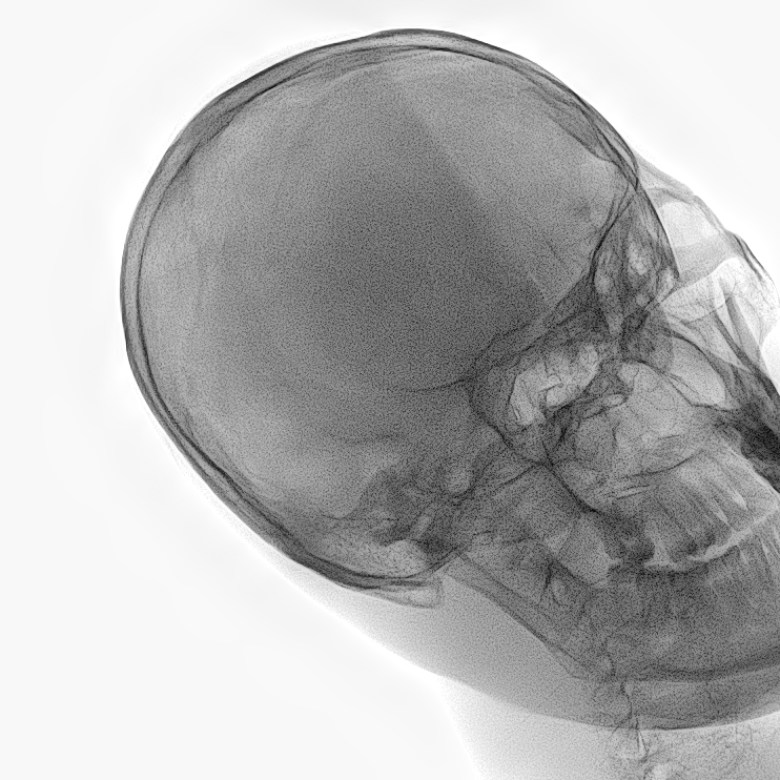} &
        \includegraphics[width=0.18\textwidth]{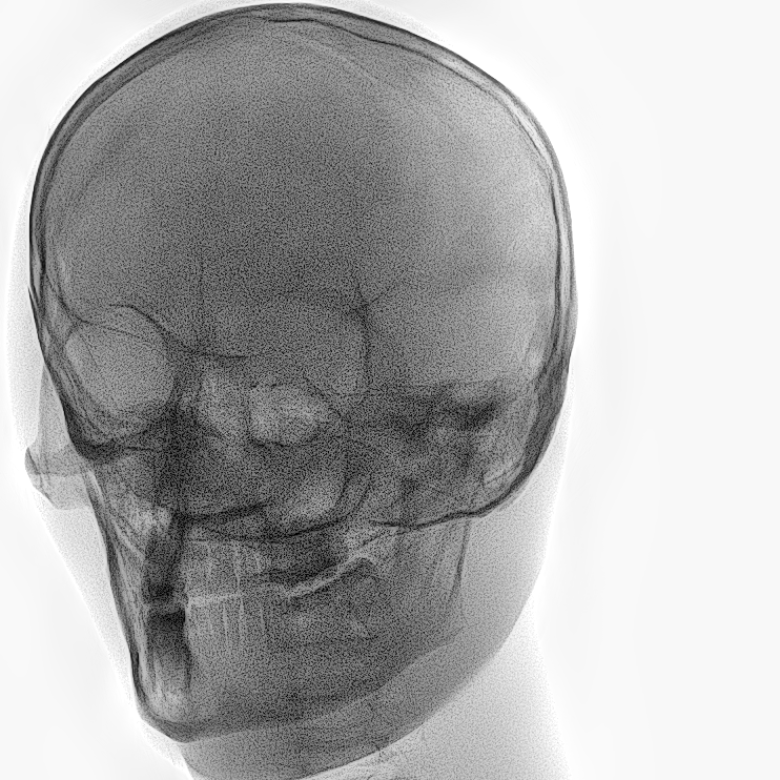} &
        \includegraphics[width=0.18\textwidth]{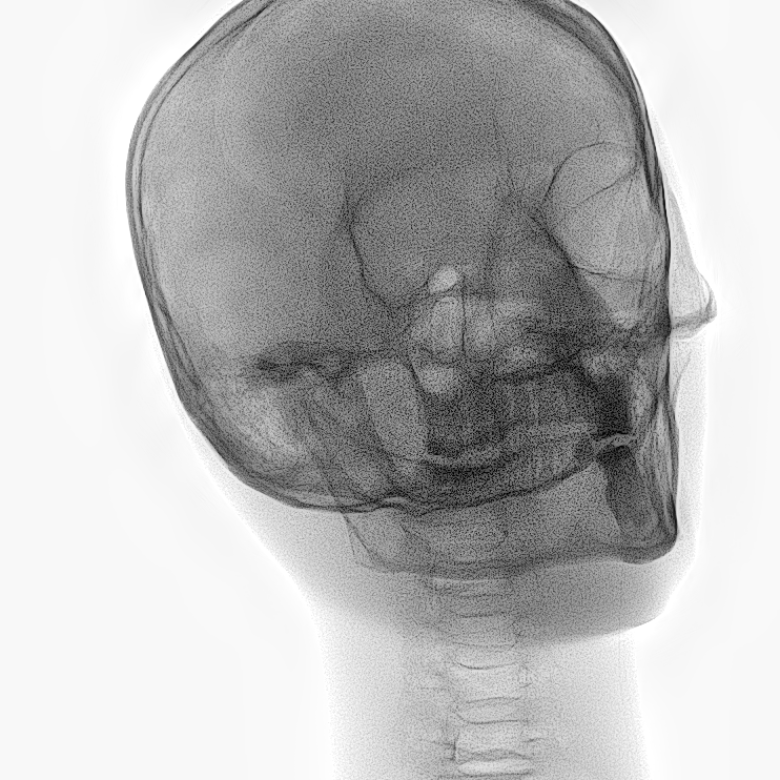}\\
        
        \raisebox{0.9cm}{MedShift~($\boldsymbol{\tau}$=0.6)} &
        \includegraphics[width=0.18\textwidth]{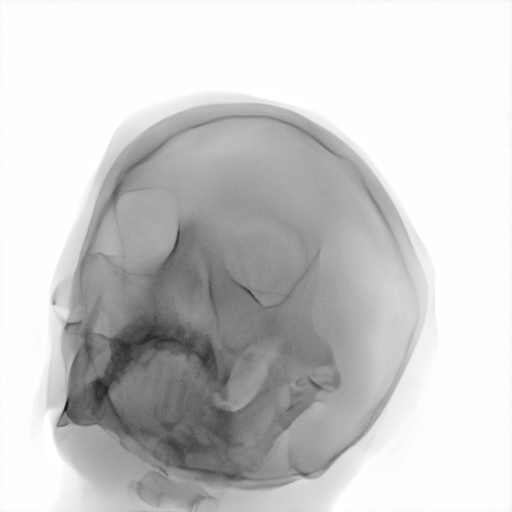} &
        \includegraphics[width=0.18\textwidth]{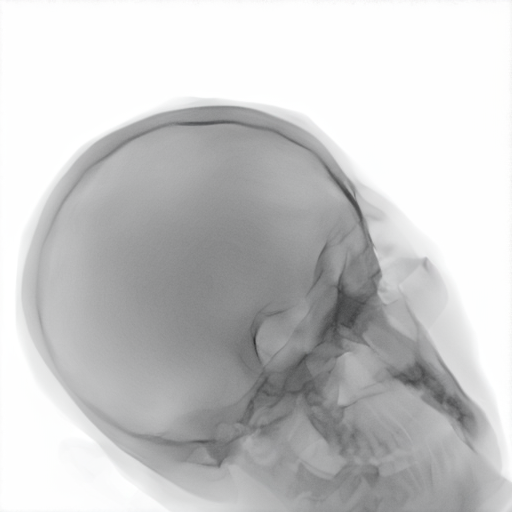} &
        \includegraphics[width=0.18\textwidth]{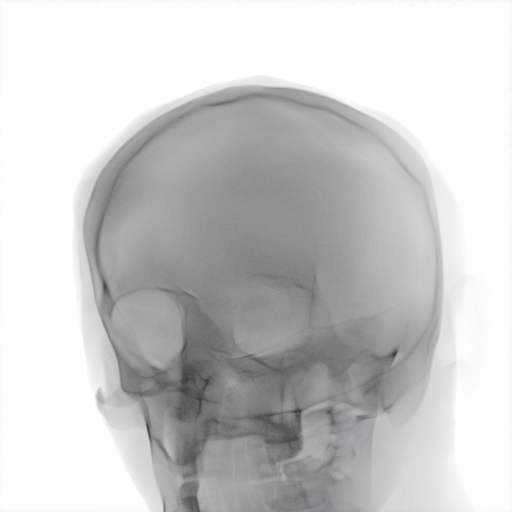} &
        \includegraphics[width=0.18\textwidth]{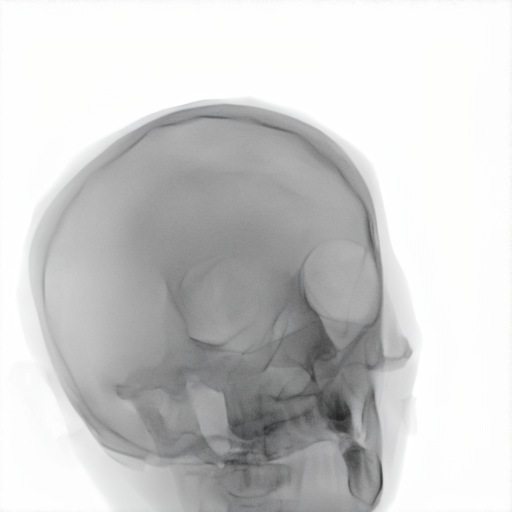} \\
        
        \raisebox{0.9cm}{MedShift~($\boldsymbol{\tau}$=0.5)} &
        \includegraphics[width=0.18\textwidth]{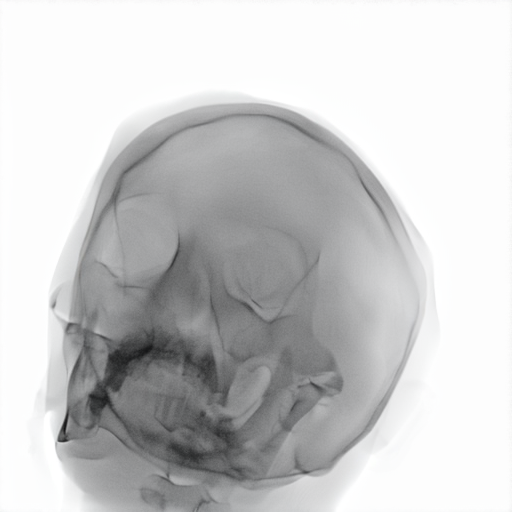} &
        \includegraphics[width=0.18\textwidth]{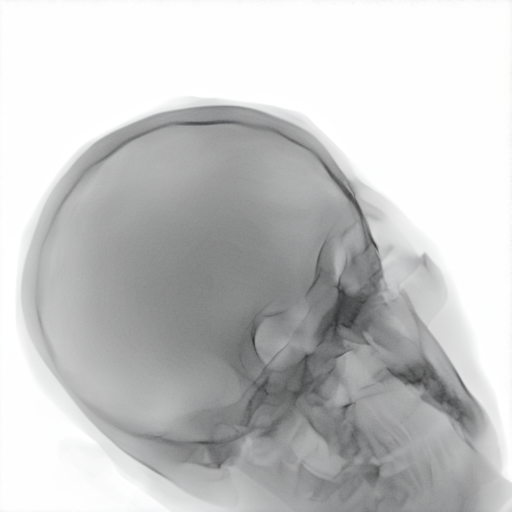} &
        \includegraphics[width=0.18\textwidth]{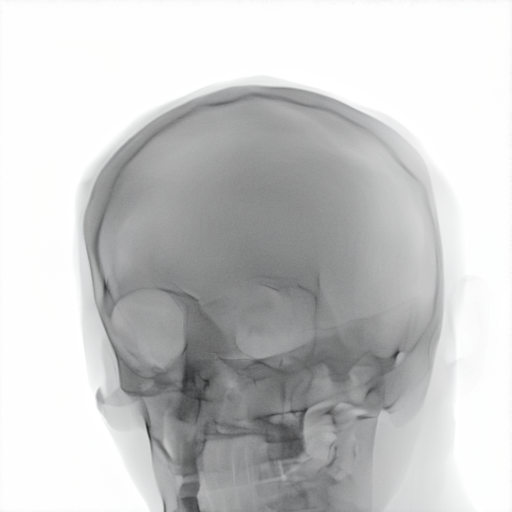} &
        \includegraphics[width=0.18\textwidth]{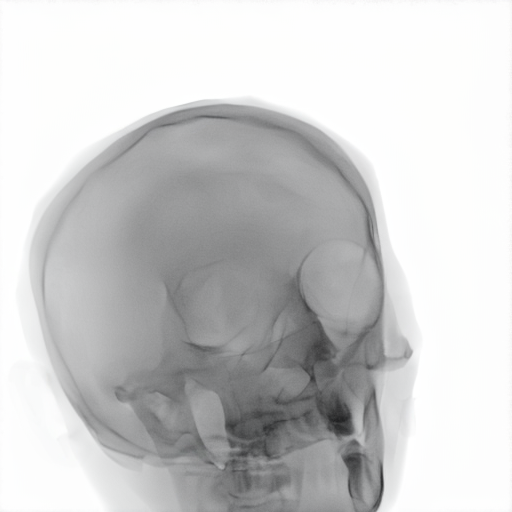} \\
        
        \raisebox{0.9cm}{MedShift~($\boldsymbol{\tau}$=0.4)} &
        \includegraphics[width=0.18\textwidth]{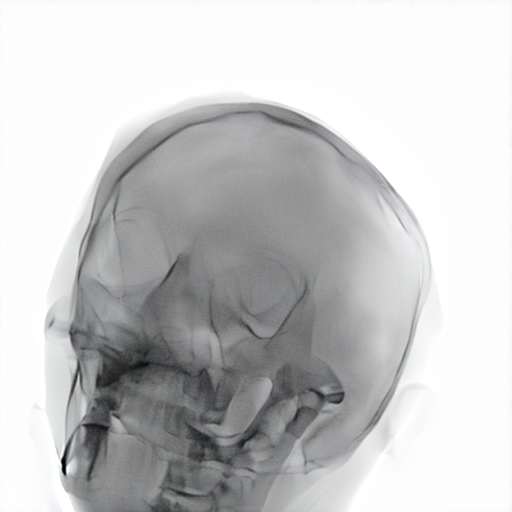} &
        \includegraphics[width=0.18\textwidth]{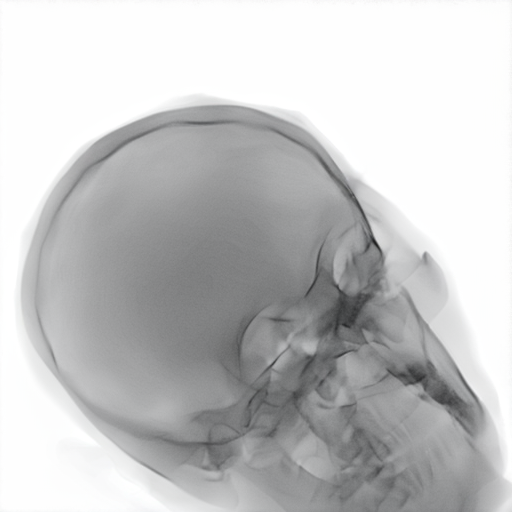} &
        \includegraphics[width=0.18\textwidth]{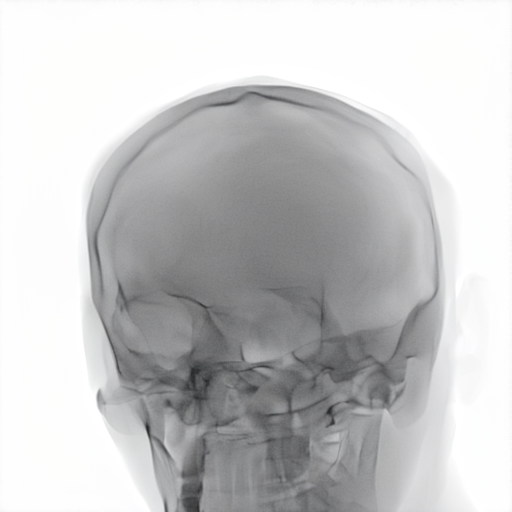} &
        \includegraphics[width=0.18\textwidth]{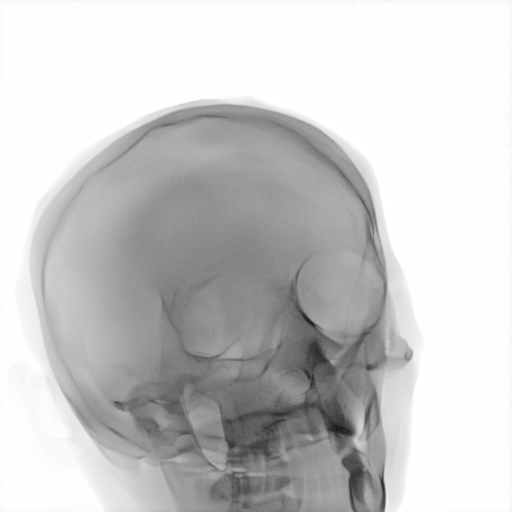} \\
        
        \raisebox{0.9cm}{MedShift~($\boldsymbol{\tau}$=0.3)} &
        \includegraphics[width=0.18\textwidth]{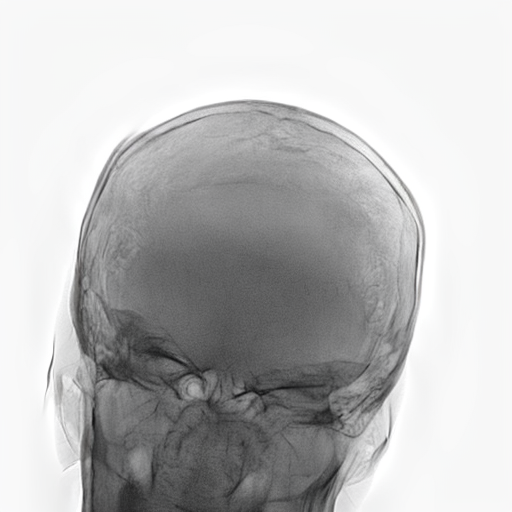} &
        \includegraphics[width=0.18\textwidth]{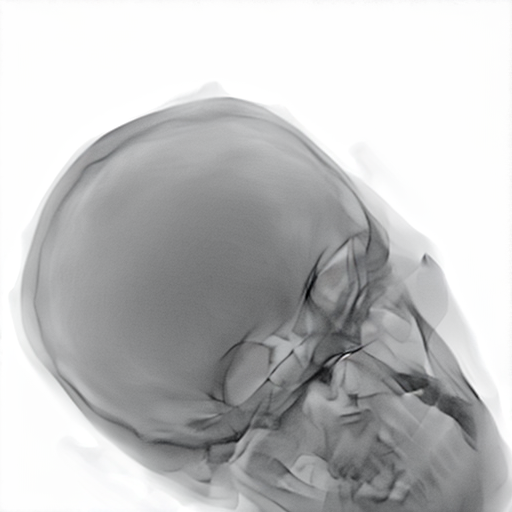} &
        \includegraphics[width=0.18\textwidth]{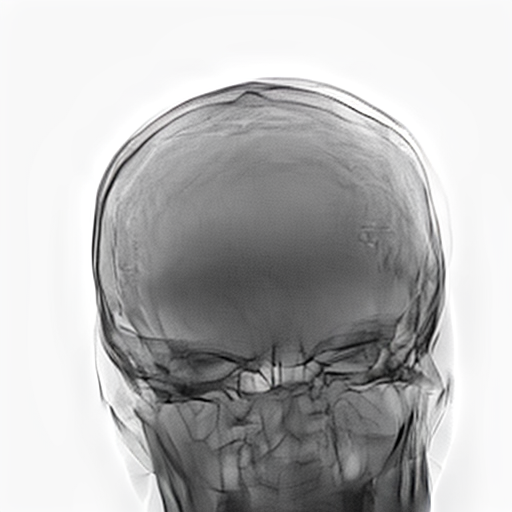} &
        \includegraphics[width=0.18\textwidth]{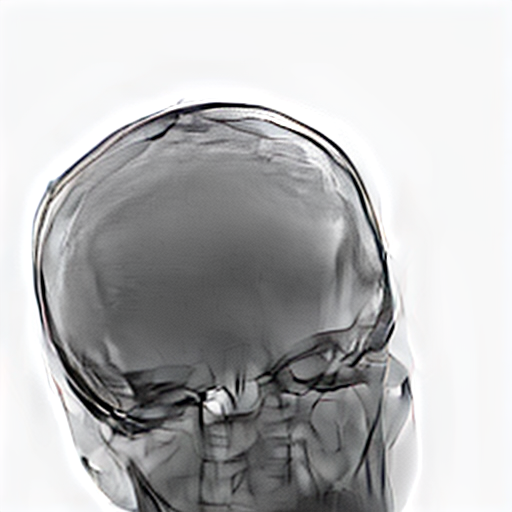} \\
        \bottomrule
    \end{tabular}
    }
    \label{tab:additional_results}
\end{table*}


\begin{table*}[t]
    \centering
    \caption{Ablation study over combinations of $\tau$ and classifier-free guidance.}
    \label{tab:ablation_tau_cfg}
    \resizebox{\textwidth}{!}{%
    \begin{tabular}{c|ccc|ccc|ccc|ccc}
        \toprule
        \multirow{2}{*}{\textbf{CFG}} & \multicolumn{3}{c|}{$\boldsymbol{\tau = 0.3}$} & \multicolumn{3}{c|}{$\boldsymbol{\tau = 0.4}$} & \multicolumn{3}{c|}{$\boldsymbol{\tau = 0.5}$} & \multicolumn{3}{c}{$\boldsymbol{\tau = 0.6}$} \\
        & \textbf{CFID($\downarrow$)} & \textbf{LPIPS($\downarrow$)} & \textbf{SSIM($\uparrow$)} & \textbf{CFID($\downarrow$)} & \textbf{LPIPS($\downarrow$)} & \textbf{SSIM($\uparrow$)} & \textbf{CFID($\downarrow$)} & \textbf{LPIPS($\downarrow$)} & \textbf{SSIM($\uparrow$)}
        & \textbf{CFID($\downarrow$)} & \textbf{LPIPS($\downarrow$)} & \textbf{SSIM($\uparrow$)}\\
        \midrule
        6.5 & 187.17 & 0.16 & 0.82 & 202.79 & 0.10 & 0.88 & 210.76 & 0.07 & 0.92 & 216.56 & \textbf{0.06} & \textbf{0.94} \\
        7.5 & 180.07 & 0.20 & 0.79 & 194.90 & 0.13 & 0.85 & 202.93 & 0.09 & 0.89 & 207.72 & 0.07 & 0.93 \\
        8.5 & 171.59 & 0.24 & 0.75 & 190.22 & 0.17 & 0.82 & 199.12 & 0.12 & 0.87 & 201.72 & 0.09 & 0.91 \\
        9.5 & \textbf{168.28} & 0.26 & 0.73 & 188.91 & 0.21 & 0.78 & 197.27 & 0.15 & 0.84 & 199.68 & 0.11 & 0.89 \\
        \bottomrule
    \end{tabular}}
\end{table*}

\subsection{Further evaluation of the translated X‐rays}

The translated radiographs in Table~\ref{tab:additional_results} replace the sharp and unnatural edges of the synthetic skull with smoother gradients that resemble the natural variation found in real X-rays. The skull contours become more realistic, showing a gradual transition at the edges rather than the artificial, razor-sharp boundaries typical of forward-rendered projections. Inside the skull, subtle intensity variations appear, following anatomical structures such as sinus cavities and internal bone texture, especially visible in Samples 3 and 4. In general, brightness is slightly reduced and contrast is improved, making the images visually closer to real clinical scans.

The model also recovers soft tissue details that are missing in the synthetic images, such as the thin scalp and the fat layer surrounding the skull. These are rendered with more realistic transitions and shading, especially near curved regions. At the boundary between bone and soft tissue, the model produces smooth transitions that mimic how real imaging systems blur the interface. Deeper regions such as the mastoid and neck base show more natural transparency and layering, leading to richer contrast and a more authentic appearance. Although the translated images exhibit markedly improved realism, some domain discrepancies persist, which may reflect both the inherent limitations of current translation methods and the need for higher-fidelity synthetic inputs.

\subsection{Computational Requirements}

We evaluate inference performance using consistent software (CUDA 12.6, PyTorch 2.6) and hardware (RTX 5090, Ryzen 9900X). Each model processes a pre-loaded validation batch using batch size 1 and FP32 precision to isolate architectural differences. Latency measurements average 100 forward passes after five warm-up steps, with each pass bracketed by \texttt{torch.cuda.synchronize()} to exclude asynchronous overhead. Peak VRAM usage is computed as the average between \texttt{torch.cuda.max\_memory\_allocated()} and NVML-reported memory, both reset before each pass. The results are reported in Table~\ref{inference_resources}.

Although Hierarchy Flow minimizes both latency and model size, its performance is suboptimal, restricting its practical applicability. CycleGAN-Turbo is a strong performer in terms of speed due to its single forward pass, but its memory footprint is prohibitively high, complicating use in local or edge scenarios. MedShift strikes a favorable balance: it is over 4 times smaller than SDEdit and CycleGAN-Turbo. This makes it a compelling option for deployment on constrained hardware. However, unlike GANs, its generative process currently involves multiple steps. Future work should focus on accelerated sampling strategies~\cite{salimans2022progressive}.

\begin{table}[!ht]
    \centering
    \caption{Inference computational requirements.}
    \label{inference_resources}
    \resizebox{0.75\linewidth}{!}{%
    \begin{tabular}{l|cc}
    \toprule
    \textbf{Model} & \textbf{Size~(MB)} & \textbf{Latency~(s)}\\
    \midrule
    Hierarchy Flow & 729.8 & 0.01 \\ 
    \midrule
    CycleGAN-Turbo & 6185.6 & 0.06\\
    \midrule
    Z-STAR & 24686.1 & 8.78 \\ 
    SDEdit~(st=0.1) & 7157.7 &  0.67 \\ 
     SDEdit~(st=0.2) & 7157.7 & 1.15 \\  
      SDEdit~(st=0.3) & 7157.7 & 1.63 \\ 
    \midrule
    MedShift~($\tau$=0.6) & 1539.5 & 0.45 \\ 
    MedShift~($\tau$=0.45) & 1539.5 & 0.62 \\ 
    MedShift~($\tau$=0.3) & 1539.5 & 0.77 \\ 
    \bottomrule
    \end{tabular}}
\end{table}

\section{Future Work}

Future work will focus on improving inference efficiency, particularly through model distillation to reduce latency while preserving output quality. Another avenue is extending the current binary translation setup to multi-class scenarios, including intra-domain mappings such as dose standardization. Additionally, incorporating auxiliary conditioning inputs-such as spatial masks or textual prompts-may enhance structural control and allow for more flexible, user-driven generation. Lightweight post-processing techniques, like histogram equalization, could also address residual issues such as low contrast. These methods should be implemented alongside the model and fine-tuned for specific deployment settings.

\section{Conclusion}

This study introduces MedShift, a flexible and unified class-conditional generative model designed for high-fidelity, unpaired image translation within the medical imaging domain. Using Flow Matching and Schrödinger bridges, MedShift effectively learns a shared domain-agnostic latent space, enabling translation between any observed training domains without requiring separate models or paired data. This work also introduces X-DigiSkull, a novel dataset comprising of synthetic and real skull X-rays acquired under varying radiation doses, establishing a robust benchmark for domain adaptation research. Our comprehensive experimental evaluation demonstrates that MedShift consistently outperforms state-of-the-art baselines, including CycleGAN-Turbo, Z-STAR, and SDEdit, across key metrics of perceptual quality and distributional alignment. MedShift achieves this with a U-Net architecture that is six times smaller than the Stable Diffusion U-Nets utilized by competing models, resulting in significantly enhanced computational efficiency. Future work includes experimenting with larger model architectures, different sampling strategies, and fine-tuning the VAE.

\section*{Acknowledgements}

The authors thank Mentice AB for providing access to the VIST\textsuperscript{\textregistered} G7 simulator and for their expert support in image-guided therapy and simulation workflows. The simulator provided the synthetic X-ray data used. We would like to acknowledge the Philips Image Guided Therapy Systems Test Automation team for their invaluable assistance with data collection. The European Xecs Eureka TASTI Project funded this research.
{
    \small
    \bibliographystyle{ieeenat_fullname}
    \bibliography{main}
}
\clearpage
\setcounter{page}{1}
\maketitlesupplementary

\appendix

The supplementary material is organized as follows: Appendix~\ref{sec:implementation_sup} describes the implementation details of MedShift. Appendix~\ref{sec:latent} contains empiric proof of the shared manifold assumption of Section~\ref{sec:medshift}.

\section{Implementation Details}
\label{sec:implementation_sup}

The model was trained on a workstation equipped with an NVIDIA RTX 3090 Ti GPU (24GB VRAM), an Intel Xeon Silver 4216 CPU (2.10 GHz), and 192GB of RAM. We used mixed-precision training via the \texttt{Accelerate}~\cite{accelerate} library to reduce memory consumption without compromising performance. The hyperparameters used to train MedShift are summarized in Table~\ref{tab:config}.

\begin{table}[htbp]
    \centering
    \caption{Model and training configuration used in our experiments.}
    \label{tab:config}
    \resizebox{0.6\linewidth}{!}{%
    \begin{tabular}{l|c}
        \toprule
        \textbf{Parameter} & \textbf{Value} \\
        \midrule
        Input size & 512 \\
        Model channels & 256 \\
        Number of residual blocks & 2 \\
        Channel multiplier & 1, 2, 2, 2 \\
        \midrule
        Attention resolutions & 2, 4 \\
        Number of attention heads & 4 \\
        Head channels & 64 \\
        \midrule
        Label Dropout probability & 0.2 \\
        Learning rate & 1e-4 \\
        Number of epochs & 1,000 \\
        Batch size & 24 \\
        Warmup steps & 100 \\
        EMA rate & 0.999 \\
        \bottomrule
    \end{tabular}}
\end{table}

\section{Latent Distributions}
\label{sec:latent}

\begin{figure}[!h]
    \centering
    \begin{subfigure}[t]{0.7\linewidth}
        \centering
        \includegraphics[width=\linewidth]{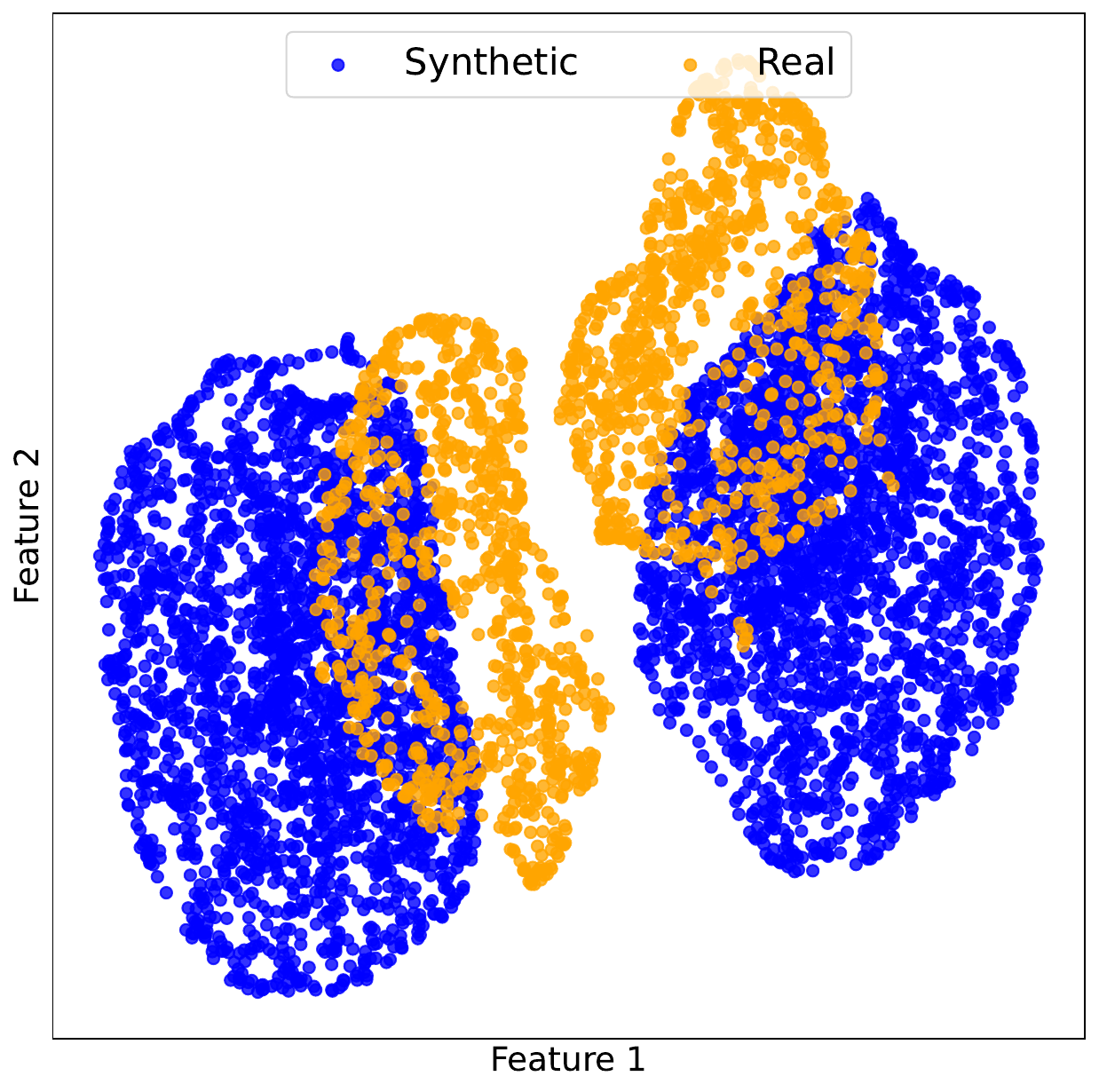}
        \caption{Original Latents~($\tau$=1.0)}
    \end{subfigure}
    \vfill
    \begin{subfigure}[t]{0.7\linewidth}
        \centering
        \includegraphics[width=\linewidth]{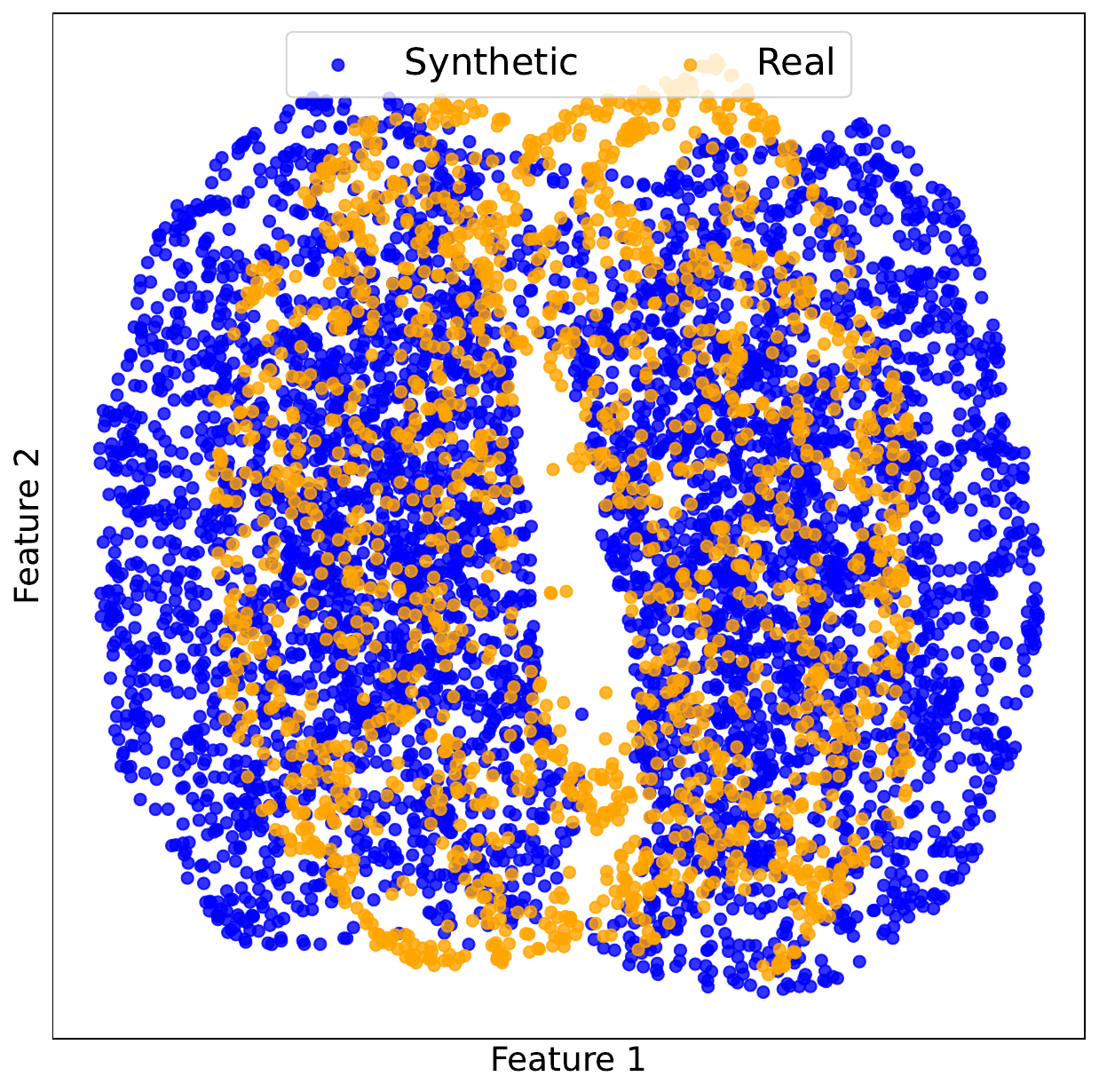}
        \caption{$\tau$=0.6}
    \end{subfigure}
    \vfill
    \begin{subfigure}[t]{0.7\linewidth}
        \centering
        \includegraphics[width=\linewidth]{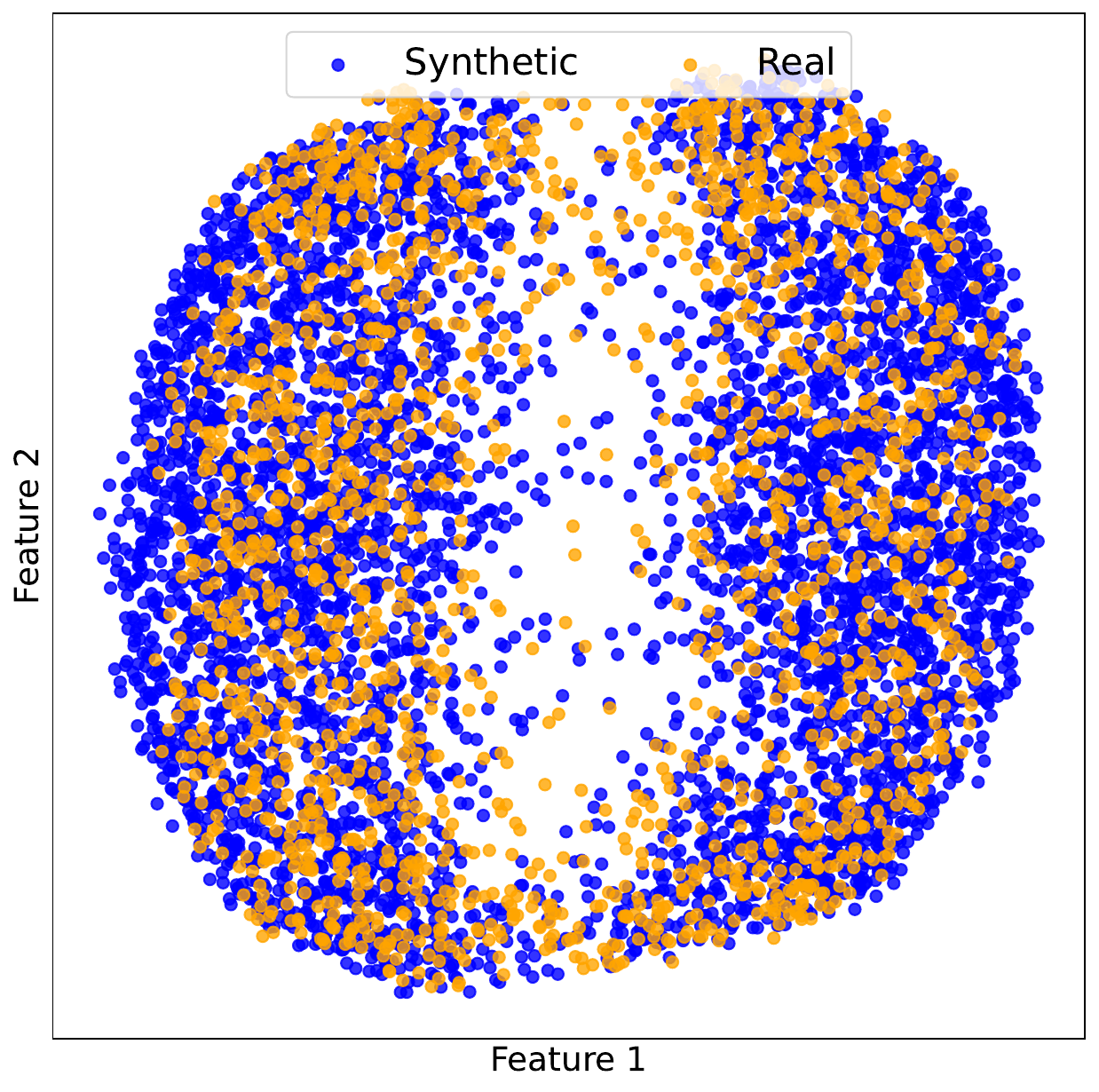}
        \caption{$\tau$=0.3}
    \end{subfigure}
    \caption{UMAP visualization of the latent-space features for different $\tau$ levels.}
    \label{fig:latents}
\end{figure}

To directly address the assumption of a shared manifold between synthetic and real domains, we add a UMAP analysis of the latent encodings for different $\tau$ values in Figure~\ref{fig:latents}. At $\tau$=1.0, where no noise is applied, the embeddings of synthetic and real images remain clearly separated. However, as $\tau$ decreases and the model integrates backward, the latent representations become progressively noisier and the two distributions begin to overlap. This analysis supports the core design of our method: by moving to an intermediate, noise-conditioned state, the model converges toward the shared latent space hypothesized in Section~\ref{sec:medshift}, enabling the subsequent domain translation.

\end{document}